\pdfoutput=1

\documentclass[11pt]{article}

\usepackage[]{emnlp2021}

\usepackage{times}
\usepackage{latexsym}

\usepackage[T1]{fontenc}

\usepackage[utf8]{inputenc}

\usepackage{microtype}

\usepackage{bm}
\usepackage{subfiles}
\usepackage{amsmath}
\usepackage{amsfonts}
\usepackage{multirow}
\usepackage[normalem]{ulem}
\usepackage{bbding}
\usepackage{CJKutf8}
\usepackage{graphicx}
\usepackage{xcolor}
\usepackage{caption}

\usepackage{subfigure}

\DeclareMathOperator{\softmax}{softmax} 
\DeclareMathOperator{\cosine}{cosine} 
\newcommand{\Japanese}[1]{{\begin{CJK}{UTF8}{min}#1\end{CJK}}}
\newcommand{\Korean}[1]{{\begin{CJK}{UTF8}{}\CJKfamily{mj}#1\end{CJK}}}

\title{A Multilingual Bag-of-Entities Model  for  \\  Zero-Shot Cross-Lingual Text Classification}

\author{
    Sosuke Nishikawa$^{1,2}$\thanks{\hspace{2mm}Work done as an intern at Studio Ousia.}\\
    {\small \texttt{sosuke-nishikawa@alumni.u-tokyo.ac.jp}}
    \And
    Ikuya Yamada$^{2,4}$\\
    {\small \texttt{ikuya@ousia.jp}}
    \AND
    Yoshimasa Tsuruoka$^{1}$\\
    {\small \texttt{tsuruoka@logos.t.u-tokyo.ac.jp}}
    \And
    Isao Echizen$^{1,3}$\\
    {\small \texttt{iechizen@nii.ac.jp}}
    \AND
    \begin{minipage}{\textwidth}
        \begin{center}
            \fontsize{11.5}{14}\selectfont
            \textnormal{$^1$The University of Tokyo\,\,\,$^2$Studio Ousia}\\
            \textnormal{$^3$National Institute of Informatics\,\,\,$^4$RIKEN AIP} \end{center}
    \end{minipage}
}

\begin{document}
\maketitle
\begin{abstract}
 We present a multilingual bag-of-entities model that effectively boosts the performance of zero-shot cross-lingual text classification by extending a multilingual pre-trained language model (e.g., M-BERT). It leverages the multilingual nature of Wikidata: entities in multiple languages representing the same concept are defined with a unique identifier. This enables entities described in multiple languages to be represented using shared embeddings.
A model trained on entity features in a resource-rich language can thus be directly applied to other languages. Our experimental results on cross-lingual topic classification (using the MLDoc and TED-CLDC datasets) and entity typing (using the SHINRA2020-ML dataset) show that the proposed model consistently outperforms state-of-the-art models.

\end{abstract}


\section {Introduction}

In the zero-shot approach to cross-lingual transfer learning, models are trained on annotated data in a resource-rich language (the source language) and then applied to another language (the target language) without any training.
Substantial progress in cross-lingual transfer learning has been made using multilingual pre-trained language models (PLMs), such as multilingual BERT (M-BERT), jointly trained on massive corpora in multiple languages \cite{devlin-etal-2019-bert, NEURIPS2019_c04c19c2, conneau-etal-2020-unsupervised}.
However, recent empirical studies have found that cross-lingual transfer learning with PLMs does not work well for languages with insufficient pre-training data or between distant languages \cite{conneau-etal-2020-emerging, lauscher-etal-2020-zero}, which suggests the difficulty of cross-lingual transfer based solely on textual information.

\begin{figure*}[t]
    \centering
    \includegraphics[width=15cm]{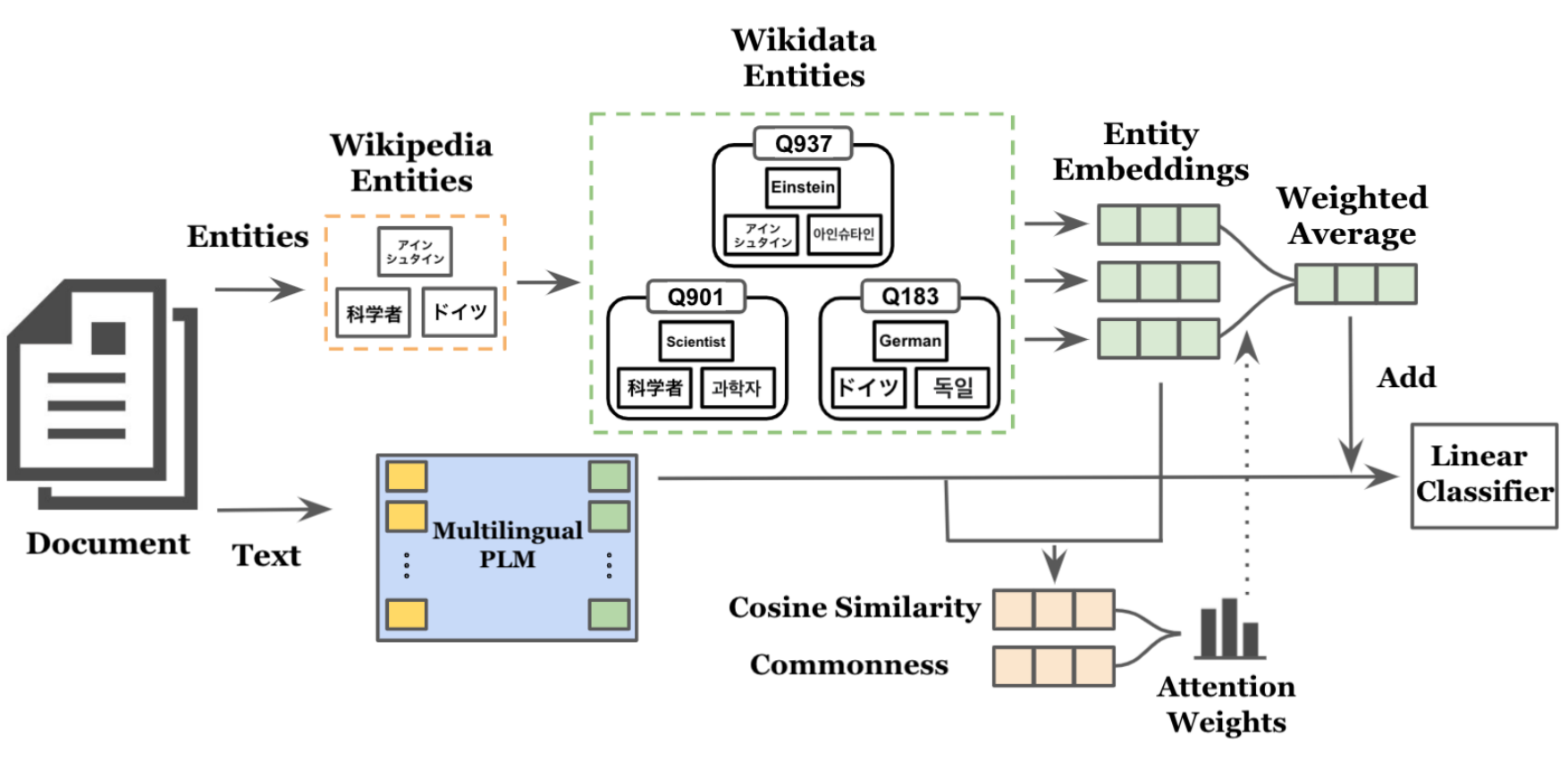}
    \caption{Architecture of M-BoE. Given a document, the model extracts Wikipedia entities, converts them into corresponding Wikidata entities, and calculates the entity-based document representation by using the weighted average of the embeddings of the entities selected by an attention mechanism. The sum of the entity-based representation and the representation computed using a multilingual PLM is used to perform linear classification for the task.}
    \label{fig:proposed_model}
\end{figure*}

We propose a multilingual bag-of-entities (M-BoE) model that boosts the performance of zero-shot cross-lingual text classification by automatically generating links to a language-agnostic knowledge base (KB) and injecting features of these entities into PLMs.
KB entities, unlike words, can capture unambiguous semantics in documents and be effectively used to address text classification tasks \cite{10.5555/1597348.1597395, 10.5555/1620163.1620201, negi-rosner-2013-uom, 10.5555/3060832.3061027, yamada-shindo-2019-neural}.
In particular, our model extends PLMs by using Wikidata entities as input features (see Figure \ref{fig:proposed_model}). 
A key idea behind our model is to leverage the multilingual nature of Wikidata: entities in multiple languages representing the same concept (e.g., \textit{Apple Inc., \Korean{애플}, \Japanese{アップル}}) are assigned a unique identifier across languages (e.g., Q312).
Given a document to be classified, our model extracts Wikipedia entities from the document, converts them into the corresponding Wikidata entities, and computes the entity-based document representation as the weighted average of the embeddings of the extracted entities. Inspired by previous work \citep{yamada-shindo-2019-neural, peters-etal-2019-knowledge}, we compute the weights using an attention mechanism that selects the entities relevant to the given document. We then compute the sum of the entity-based document representation and the text-based document representation computed using the PLM and feed it into a linear classifier.
Since the entity vocabulary and entity embedding are shared across languages, a model trained on entity features in the source language can be directly transferred to multiple target languages.


We evaluate the performance of the M-BoE model on three cross-lingual text classification tasks: topic classification on the MLDoc \cite{schwenk-li-2018-corpus} and TED-CLDC \cite{hermann-blunsom-2014-multilingual} datasets and entity typing on the SHINRA2020-ML \cite{mioverview} dataset.
We train the model using training data in the source language (English) and then evaluate it on the target languages.
It outperforms our base PLMs (i.e., M-BERT \cite{devlin-etal-2019-bert} and the XLM-R model \cite{conneau-etal-2020-unsupervised}) for all target languages on all three tasks, thereby demonstrating the effectiveness of the entity-based representation.
Furthermore, our model performs better than state-of-the-art models on the MLDoc dataset.

Our contributions are as follows:
\begin{itemize}
  \item We present a method for boosting the performance of cross-lingual text classification by extending multilingual PLMs to leverage the multilingual nature of Wikidata entities. 
Our method successfully improves the performance on multiple target languages simultaneously without expensive pre-training or additional text data in the target languages.

  \item Inspired by previous work \cite{yamada-shindo-2019-neural, peters-etal-2019-knowledge}, we introduce an attention mechanism that enables entity-based representations to be effectively transferred from the source language to the target languages. The mechanism selects entities that are relevant to address the task.
  
  \item We present experimental results for three cross-lingual text classification tasks demonstrating that our method outperforms our base PLMs (i.e., M-BERT and XLM-R) for all languages on the three tasks and outperforms state-of-the-art methods on the MLDoc dataset.
\end{itemize}


\section {Related Work}

\label{sec: relatedwork}

\paragraph{Cross-lingual PLMs}
Zero-shot cross-lingual transfer learning approaches have relied on parallel corpora \citep{xu-wan-2017-towards} or multilingual word representation \citep{duong-etal-2017-multilingual}.
Considerable progress has been made on PLMs for various cross-lingual transfer tasks. 
The representative models are M-BERT \citep{devlin-etal-2019-bert} and XLM-R \citep{conneau-etal-2020-unsupervised}, which are multilingual extensions of BERT \cite{devlin-etal-2019-bert} and RoBERTa \cite{DBLP:journals/corr/abs-1907-11692}, respectively. Both models are pre-trained on massive corpora of approximately 100 languages.
LASER \citep{DBLP:journals/tacl/ArtetxeS19} is a PLM trained on a parallel corpus of 93 languages by using a sequence-to-sequence architecture.

\paragraph{Improving cross-lingual transfer learning }
Several studies have attempted to improve cross-lingual transfer learning by using additional text data in the target language.
\citet{DBLP:journals/corr/abs-1909-07009} proposed using an unlabeled corpus in the target language to bridge the gap between the language and the domain.
\citet{10.1145/3397271.3401209} and \citet{keung-etal-2019-adversarial} incorporated adversarial training using unlabeled target language examples.
\citet{dong-de-melo-2019-robust} and \citet{eisenschlos-etal-2019-multifit} presented methods for data augmentation in which pseudo-labels are assigned to an unlabeled corpus in the target language.
\citet{NEURIPS2019_c04c19c2} additionally pre-trained BERT-based models using a parallel corpus.
However, these methods require extra training on additional text data for each target language, and their resulting models work well only on a single target language. Unlike these methods, our method does not require extra training and improves performance simultaneously for all target languages with only a single PLM. Furthermore, our method can be easily applied to these models since it is a simple extension of a PLM and does not modify its internal architecture.

\paragraph{Enhancing monolingual PLMs using entities } 
Several methods have been proposed for improving the performance of PLMs through pre-training using entities.
ERNIE \citep{zhang-etal-2019-ernie} and KnowBert \citep{peters-etal-2019-knowledge} enrich PLMs by using pre-trained entity embeddings. LUKE \citep{yamada-etal-2020-luke} and EaE \citep{fevry-etal-2020-entities} train entity embeddings from scratch during pre-training.
However, all of these methods are aimed at improving the performance of monolingual tasks and require pre-training with a large corpus, which is computationally expensive.
Our method dynamically injects entity information into PLMs during fine-tuning without expensive pre-training.

Several studies have attempted to incorporate entity information into PLMs after pre-training to enhance the performance of monolingual tasks. 
\citet{Ostendorff2019} concatenated contextualized representations with knowledge graph embeddings to represent author entities and used them as features for the book classification task. 
E-BERT \cite{poerner-etal-2020-e} inserts KB entities next to the entity names in the input sequence to improve BERT's performance for entity-centric tasks.
\citet{verlinden-etal-2021-injecting} introduced a mechanism for combining span representations and KB entity representations within a BiLSTM-based end-to-end information extraction model.
Unlike these methods, our method aims to improve the cross-lingual text classification by combining PLMs with language-agnostic entity embeddings.

\paragraph{Text classification models using entities}
Several methods have been commonly used to address text classification using entities.
Explicit semantic analysis (ESA) is a representative example; it represents a document as a bag of entities, which is a sparse vector in which each dimension is a score reflecting the relevance of the text to each entity \cite{10.5555/1597348.1597395, 10.5555/1620163.1620201, negi-rosner-2013-uom}.
More recently, \citet{10.5555/3060832.3061027} proposed cross-lingual explicit semantic analysis (CLESA), an extension of ESA, to address cross-lingual text classification. CLESA computes sparse vectors from the intersection of Wikipedia entities in the source and target languages using Wikipedia language links.
Unlike CLESA's approach, we address cross-lingual text classification by extending state-of-the-art PLMs with a language-agnostic entity-based document representation based on Wikidata.

The most relevant to our proposed approach is the neural attentive bag-of-entities (NABoE) model proposed by \citet{yamada-shindo-2019-neural}.
It addresses monolingual text classification using entities as inputs and uses an attention mechanism to detect relevant entities in the input document.
Our model can be regarded as an extension of NABoE by (1) representing documents using a shared entity embedding across languages and (2) combining an entity-based representation and attention mechanism with state-of-the-art PLMs.


\section{Proposed Method}
\label{sec: approach}

\begin{table*}[t]
\centering
 \scalebox{0.8}{ 
\begin{tabular}{ccrrrrr}
\hline
\  Dataset & Language  & Train & Dev. & Test\\
\hline
MLDoc  &  8     & 1,000 & 1,000 & 4,000 \\

TED-CLDC  &  12     & 936 & 105 & 51--106 \\

SHINRA &  30    & 417,387 &21,967 & 30k--920k  \\
\hline
\end{tabular}
}
\caption{Number of examples in MLDoc, TED-CLDC, and SHINRA2020-ML datasets.}
\label{tab:dataset}
\end{table*}

Figure \ref{fig:proposed_model} shows the architecture of our model.
The model extracts Wikipedia entities, converts them into Wikidata entities, and computes the entity-based document representation using an attention mechanism.
The sum of the entity-based document representation and the text-based document representation computed using the PLM is fed into a linear classifier to perform classification tasks.

\subsection{Entity detection}
\label{subsec:entity}

To detect entities in the input document, we use two dictionaries that can be easily constructed from the KB: (1) a mention-entity dictionary, which binds an entity name (e.g., ``Apple'') to possible referent KB entities (e.g., \textit{Apple Inc.} and \textit{Apple (food))} by using the internal anchor links in Wikipedia \cite{guo-etal-2013-link}, and (2) an inter-language entity dictionary, which links multilingual Wikipedia entities (e.g., \textit{Tokyo, \Korean{도쿄}, \Japanese{東京}}) to a corresponding identifier (e.g., Q7473516) of Wikidata.

All words and phrases are extracted from the given document in accordance with the mention-entity dictionary\footnote{Following past work~\cite{yamada-shindo-2019-neural}, name overlap bounds are resolved by detecting only the earliest and longest ones.}, and all possible referent entities are detected if they are included as entity names in the dictionary.
Note that all possible referent entities are detected for each entity name rather than a single resolved entity. For example, we detect both \textit{Apple Inc.} and \textit{Apple (food)} for entity name ``Apple''.
Next, the detected entities are converted into Wikidata entities if they are included in the inter-language entity dictionary.

\subsection{Model}
\label{subsec:attention}

Each Wikidata entity is assigned a representation $\boldsymbol{v}_{e_i} \in \mathbb{R}^d$.
Since our method extracts all possible referent entities rather than a single resolved entity, it often extracts entities that are not related to the document. Therefore, we introduce an attention mechanism inspired by previous work \citep{yamada-shindo-2019-neural, peters-etal-2019-knowledge} to prioritize entities related to the document.
 Given a document with $K$ detected entities, our method computes the entity-based document representation $  \boldsymbol{z} \in \mathbb{R}^d$ as the weighted average of the entity embeddings:

\begin{equation}
\boldsymbol{z} = \sum_{i=1}^K a_{e_{i}} \boldsymbol{v}_{e_{i}},
\end{equation}

\noindent where $a_{e_{i}} \in \mathbb{R} $ is the attention weight corresponding to entity $e_{i}$ and calculated using
\begin{gather}
\boldsymbol{a} = \softmax (\boldsymbol{W}_{a}^\top \boldsymbol{\phi}),\\
  \phi(e_{i}, d)  = \left[
       \begin{array}{c}
       \cosine (\boldsymbol{h}, \boldsymbol{v}_{e_{i}}) \\
       p_{e_{i}}
       \end{array}
  \right]
\end{gather}
\noindent where $\boldsymbol{a} = [a_{e_1}, a_{e_2},\cdots, a_{e_K}]$ are the attention weights; $\boldsymbol{W}_{a}  \in \mathbb{R}^2$ is a weight vector;
$\boldsymbol{\phi}= [\phi(e_1, d), \phi(e_2, d),\cdots , \phi(e_K, d)] \in \mathbb{R}^{2 \times K}$ represents the degree to which each entity $e_{i}$ is related to document $d$; and $\phi(e_{i}, d)$ is calculated by concatenating commonness\footnote{Commonness \cite{Mihalcea:2007:WLD:1321440.1321475} is the probability that an entity name refers to an entity in Wikipedia.} $p_{e_{i}}$ with the cosine similarity between the document representation computed using the PLM, $\boldsymbol{h} \in \mathbb{R}^d$ (e.g., the final hidden state of the [CLS] token), and entity embedding, $\boldsymbol{v}_{e_{i}}$.

The sum of this entity-based document representation  $  \boldsymbol{z}$ and text-based document representation $\boldsymbol{h}$ is fed into a linear classifier\footnote{In preliminary experiments, we also tested concatenation, but observed worse overall results than with summation. } to predict the probability of label $c$:

\begin{equation}
p(c\mid \boldsymbol{h}, \boldsymbol{z}) = Classifier(\boldsymbol{h} +\boldsymbol{z}).
\end{equation}


\section {Experimental Setup}

In this section, we describe the experimental setup we used for the three cross-lingual text classification tasks.


\begin{table*}[t]
\centering
 \scalebox{0.90}{ 
\begin{tabular}{l|l|ccccccc|cc}
\hline   Model  &  en  & fr & de & ja & zh & it & ru & es & target avg.\\
\hline 
MultiCCA \cite{schwenk-li-2018-corpus}  & 92.2 & 72.4 & 81.2 & 67.6 & 74.7 & 69.4& 60.8 &  72.5 & 71.2\\ 
LASER \cite{DBLP:journals/tacl/ArtetxeS19}  & 89.9 & 78.0 & 84.8 & 60.3 & 71.9 & 69.4 & 67.8 &  77.3 & 72.8\\ 
M-BERT   & 94.0 & 79.4 & 75.1 & 69.3 & 68.0 & 67.1& 65.3 &  75.2 & 71.4 $\pm$ 1.4\\
\qquad +M-BoE & 94.1 & 84.0 & 76.9 & 71.1& 72.2 & 70.0 & 68.9 &  75.5 & 74.1 $\pm$ 0.7 \\
 XLM-R  & 94.4 & 84.9 & 86.7 & 78.5 & 85.2 & 73.4 & 71.3 &  81.5 & 80.2 $\pm$ 0.5\\
 \qquad +M-BoE & \textbf{94.6} & \textbf{86.4} & \textbf{88.9} & \textbf{80.0}& \textbf{87.4} & \textbf{75.6} & \textbf{73.7} &  \textbf{83.2} & \textbf{82.2} $\pm$ \textbf{0.6}  \\
\hline
\end{tabular}
}
\caption{Classification accuracy for topic classification on MLDoc dataset; ``target avg.'' indicates average scores for target languages.}
\label{tab:result_mldoc}
\end{table*}



\begin{table*}[t]
\centering
 \scalebox{0.85}{ 
\begin{tabular}{l|l|ccccccccccc|c}
\hline
\   Model   &en  & fr & de & it & ru & es & ar & tr & nl & pt & pl & ro & target avg. \\ \hline 
M-BERT &51.6 & 47.7 & 43.9 & 50.6 & 47.9  & 53.1 & 41.3  &  44.2 & 49.4 & 46.2 & 45.1 & 45.4 & 47.1 $\pm$ 1.4 \\
\qquad +M-BoE & \textbf{52.9} & 49.5 & 46.2 & \textbf{53.3} & 49.2& \textbf{54.7} & 44.7 & 49.1 &51.0 & 47.6 & 47.7 & 48.2 & 49.6 $\pm$ 1.1\\ 
XLM-R & 51.5 & 49.5 & 49.7 & 48.7 & 48.3  & 51.2 & 45.6  &  51.3 & 48.8 & 46.3 & 48.3  & 48.4 & 49.1 $\pm$ 1.8 \\
\qquad +M-BoE  & 51.7 & \textbf{50.0} & \textbf{53.8} & 51.3 & \textbf{52.3}& 52.9 & \textbf{50.5} &  \textbf{53.1} & \textbf{52.0} & \textbf{49.3} &\textbf{50.5} & \textbf{49.6} & \textbf{51.8} $\pm$ \textbf{0.9}\\ \hline
\end{tabular}
}
 \caption{F1 score for topic classification on TED-CLDC dataset.}
\label{tab:result_ted}
\end{table*}


\begin{table*}[t] 
\centering
 \scalebox{0.72}{ 
\begin{tabular}{l|ccccccccccccccl}
\hline
\    &fr & de & ja  & zh & it & ru & es & ar & tr & nl & pt & pl & ro  & hi & \hspace{0.2em} no \\
\hline 
M-BERT &68.5&84.2&81.3&80.7&85.2&81.4&85.6&57.4&50.7&55.6&80.4&77.7&76.9&81.8  & 83.6  \\
\qquad +M-BoE & 69.3&\textbf{85.1}&\textbf{82.5}&\textbf{82.2}&\textbf{86.4}&\textbf{83.2}&\textbf{86.6}&61.9&54.0&59.0&\textbf{81.7}&79.4&\textbf{80.5}&82.9&\textbf{84.8}\\ 
\qquad +Oracle M-BOE & 75.4&85.2&81.9 &81.8&86.5&83.0&86.5&61.9&53.7&61.7&81.8&79.7&79.9&83.0& 84.8 \\
XLM-R & 73.0&82.6&77.4&75.1&84.2&81.0&85.3&58.9&69.1&63.7&79.8&80.0&76.9&83.3& 82.4  \\
\qquad +M-BoE & \textbf{77.4}&84.5&79.0&77.0&85.6&\textbf{83.2}&85.8&\textbf{63.3}&\textbf{72.3}&\textbf{65.5}&80.7&\textbf{81.8}&77.8&\textbf{84.8}&84.0 \\
\qquad +Oracle M-BOE & 76.5&84.8&79.6&77.2&85.5&83.4&86.2&63.0&71.8&67.6&80.4&81.5&78.8&84.8&83.2 \\
\hline

\   & th  & ca & da & fa & id & sv & vi & bg & cs & fi & he & hu  & ko & uk & target avg.   \\ \hline 
M-BERT &84.0&81.5&80.1&80.2&72.4&79.4&79.3&74.0&74.6&75.7&74.0&77.1&81.3&78.0 & 76.6 $\pm$ 0.7   \\
\qquad +M-BoE &\textbf{85.1}&\textbf{83.2}&81.4&82.1&75.4&\textbf{82.4}&81.2&76.1&76.8&\textbf{77.6}&\textbf{78.1}&\textbf{79.2}&82.9&\textbf{80.0}  & 78.7 $\pm$ 0.5  \\ 
\qquad +Oracle M-BOE &85.3&83.2&82.3&82.4&75.5&82.0&81.6&76.6&77.4&77.4&77.8&78.7&83.3&79.9 & 79.0 $\pm$ 0.5  \\
XLM-R & 81.4&79.0&81.0&82.4&75.5&75.5&80.7&76.0&77.9&74.7&70.5&73.1&82.6&74.3& 77.1 $\pm$ 1.2   \\
\qquad +M-BoE  & 82.1&80.9&\textbf{83.3}&\textbf{84.1}&\textbf{78.2}&78.7&\textbf{81.9}&\textbf{79.1}&\textbf{79.6}&76.9&71.9&75.5&\textbf{84.0}&77.0 &  \textbf{79.2} $\pm$ \textbf{0.9}  \\
\qquad +Oracle M-BOE&81.8&81.2&82.9&83.9&78.3&78.2&82.5&79.1&79.9&77.1&71.8&75.8&83.92&76.9 & 79.2 $\pm$ 0.9\\
\hline

\end{tabular}
}
 \caption{F1 score for entity typing on SHINRA2020-ML dataset.}
\label{tab:result_shinra}
\end{table*}

\label{sec: experiments}
\subsection{Entity preprocessing}

We constructed a mention-entity dictionary from the January 2019 version of Wikipedia dump\footnote{\url{https://dumps.wikimedia.org/}} and an inter-language entity dictionary from the March 2020 version in the Wikidata dump,\footnote{\url{https://dumps.wikimedia.org/wikidatawiki/entities/}} which contains 45,412,720 Wikidata entities (e.g., Q312).
We computed the commonness values from the same versions of Wikipedia dumps in the corresponding language, following the work of \citet{yamada-shindo-2019-neural}.

We initialized Wikidata entity embeddings using pre-trained English entity embeddings trained on the KB.
To train these embeddings, we used the open-source Wikipedia2Vec tool \cite{yamada-etal-2020-wikipedia2vec}.
We used the January 2019 English Wikipedia dump mentioned above and set the dimension to 768 and the other parameters to the default values.
We initialized an entity embedding using a random vector if the entity did not exist in the Wikipedia2Vec embeddings. 
Note that we used only English Wikipedia to train the entity embeddings.

\subsection{Data} 
We evaluated our model using three datasets: MLDoc \cite{schwenk-li-2018-corpus}, TED-CLDC \cite{hermann-blunsom-2014-multilingual}, and SHINRA2020-ML \cite{mioverview}.

\paragraph{MLDoc} is a dataset for multi-class text classification, i.e., classifying news articles into four categories in eight languages. We used the english.train.1000 and english.dev datasets, which contain 1000 documents for training and validation data.
As in the previous work \cite{schwenk-li-2018-corpus,keung-etal-2020-dont}, we used accuracy as the metric. 

\paragraph{TED-CLDC} is a multi-label classification dataset covering 15 topics in 12 languages based on the transcripts of TED talks. This topic classification dataset is exactly like the MLDoc dataset except that the classification task is more difficult because of its colloquial nature and because the amount of training data is small.
Following the previous work \cite{hermann-blunsom-2014-multilingual}, we used micro-average F1 as the metric. 

\paragraph{SHINRA2020-ML} is an entity typing dataset that assigns fine-grained entity labels (e.g., Person, Country, Government) to a Wikipedia page.
We used this dataset for multi-label classification tasks; we used all datasets in 30 languages except English for the test data.
Note that our model does not use information in the test data during training because we only use the English Wikipedia to train our entity embeddings.
Following the original work \cite{mioverview}, we used micro-average F1 as the metric.

We created a validation set by randomly selecting 5\% of the training data in TED-CLDC and 5\% of the training data in SHINRA2020-ML. In all experiments, we trained our model on English training data, optimized hyper-parameters using English development data, and evaluated it on the remaining languages.
A summary of the datasets is shown in Table \ref{tab:dataset}.


\subsection{Models} 
We used M-BERT \cite{devlin-etal-2019-bert} and XLM-R$_{base}$ \cite{conneau-etal-2020-unsupervised} as the baseline multilingual PLMs to evaluate the proposed method.
We added a single fully-connected layer on top of the PLMs and used the final hidden state $\boldsymbol{h}$ of the first [CLS] token as the text-based document representation.
For the MLDoc dataset, we trained the model by minimizing the cross-entropy loss with softmax activation.
For the TED-CLDC and SHINRA2020-ML datasets, we trained the model by minimizing the binary cross-entropy loss with sigmoid activation. For these two tasks, we regarded each label as positive if its corresponding predicted probability was greater than 0.5 during inference.

For topic classification using MLDoc, we compared the performance of the proposed model with those of two state-of-the-art cross-lingual models: LASER  \cite{DBLP:journals/tacl/ArtetxeS19} (see Section \ref{sec: relatedwork}), and MultiCCA \cite{schwenk-li-2018-corpus}, which is based on a convolutional neural network with multilingual word embeddings.
To ensure a fair comparison, we did not include models that use additional unlabeled text data or a parallel corpus to train models for each target language.

For entity typing, we tested a model that uses oracle entity annotations (i.e., hyperlinks) contained in the Wikipedia page to be classified instead of entities detected using the entity detection method described in Section \ref{subsec:entity}.
Note that this model also uses attention mechanisms and pre-trained entity embeddings.

\subsection{Detailed settings}


We tuned the hyper-parameters on the basis of the English validation set. 
The details on the hyperparameters of the models can be found in Appendix \ref{appendix:hyper-parameter_details}.
We trained the models using the AdamW optimizer with a gradient clipping of $1.0$.

In all experiments, we trained the models until the performance on the English validation set converged.
We conducted all experiments ten times with different random seeds, and recorded the average scores and 95\% confidence intervals.


\section{Results}
\label{sec: results}

Tables \ref{tab:result_mldoc}, \ref{tab:result_ted}, and \ref{tab:result_shinra} show the results of our experiments.
Overall, the M-BoE models outperformed their baselines (i.e., M-BERT and XLM-R) for all target languages on all three datasets.
Furthermore, there was a significant difference in the mean scores for the target languages for those models in a paired t-test ($p <0.05$).
In particular, the performance of our model clearly exceeded that of the M-BERT baseline by 2.7\% in accuracy, 2.5\% in F1, and 2.1\% in F1, on the MLDoc, TED-CLDC, and SHINRA2020-ML datasets, respectively.

For entity typing, using the entities detected with our simple dictionary-based approach achieved comparable performance to using gold entity annotations (Table \ref{tab:result_shinra}: Oracle M-BoE) on the SHINRA2020-ML dataset, which clearly demonstrates the effectiveness of our attention-based entity detection method.


\begin{table}[t]
\centering
 \scalebox{0.85}{ 
\begin{tabular}{l|cc}
\hline
\multirow{3}{*}{Setting} & M-BoE & M-BoE \\
 &  (M-BERT) & (XLM-R)  \\
 & target avg. & target avg. \\
\hline
Full model  & \textbf{74.1}  & \textbf{82.2} \\
\hline
\textbf{Attention mechanism:} \\
\qquad without attention   & 70.5  & 81.1   \\
\qquad commonness only & 72.4 & 81.8   \\
\qquad cosine only & 72.8 & 81.8  \\
\hline
\textbf{Entity embeddings:} \\
\qquad random vectors    & 73.0 & 80.9  \\
\qquad KG embedding     & 73.2 & 81.4  \\
\hline
\textbf{Entity detection method:} \\
\qquad entity linking     & 71.7 & 80.5  \\
\qquad entity linking + att   & 73.0 & 81.9  \\
\hline
Baseline  & 71.4  & 80.2 \\
\hline
\end{tabular}
}
\caption{Results of analysis of our model on MLDoc.}
\label{tab:result_analysis}
\end{table}

\begin{table*}[t]
\centering
 \scalebox{0.95}{ 
\begin{tabular}{l|r|ccccccc|cc}
\hline   Model  &  en (train)  & fr & de & ja & zh & it & ru & es & avg.\\
\hline
External entity linking & 20.0 & 19.2 & 14.6 & 8.15 &  5.2 & 11.7 & 12.7 & 13.8 & 13.2  \\
Dictionary-based method (ours)  & 105.8 & 97.8 & 78.9 & 47.9 & 34.5 & 53.2 & 64.6 & 72.3& 64.2  \\
\hline
\end{tabular}
}
\caption{Comparison of the number of detected entities on MLDoc dataset. Numbers indicate average number of entities detected for each example.}
\label{tab:detection_method}
\end{table*}

\begin{figure*}[htbp]
 \begin{minipage}{0.5\hsize}
  \begin{center}
  \includegraphics[width=8.5cm]{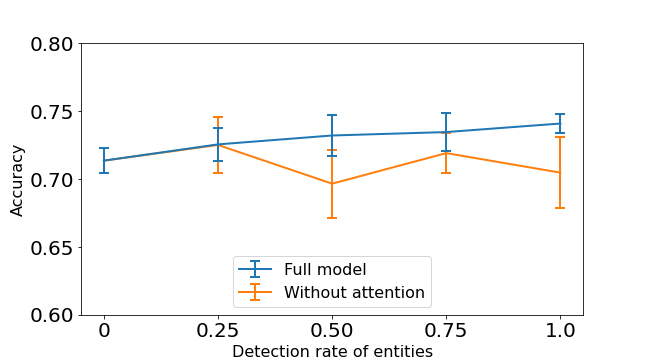}
  \end{center}
  \centering{(a) M-BoE (M-BERT)}
 \end{minipage}
 \begin{minipage}{0.5\hsize}
  \begin{center}
  \includegraphics[width=8.5cm]{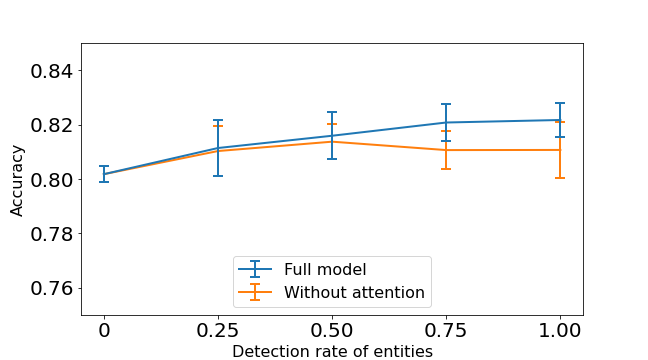}
  \end{center}
  \centering{(b) M-BoE (XLM-R)}
  \captionsetup[figure]{labelformat=empty,labelsep=none}
 \end{minipage}
 \caption{Classification accuracy for each entity detection rate using MLDoc dataset. }
 \label{fig:entity_ratio}
\end{figure*}

  
  \begin{figure*}[t]
 \begin{minipage}{0.5\hsize}
  \begin{center}
  \includegraphics[width=8.5cm]{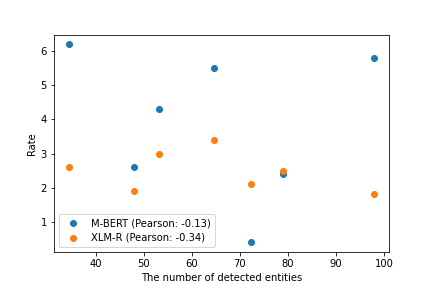}
  \end{center}
  \centering{(a) MLDoc}
 \end{minipage}
 \begin{minipage}{0.5\hsize}
  \begin{center}
  \includegraphics[width=8.5cm]{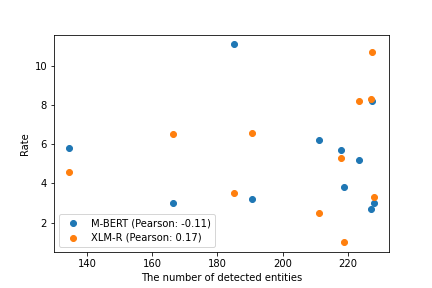}
  \end{center}
  \centering{(b) TED-CLDC}
  \captionsetup[figure]{labelformat=empty,labelsep=none}
 \end{minipage}
 \caption{Pearson correlation coefficient and scatter plot of average number of detected entities and rate of improvement in performance (Rate) for each target language.}
 \label{fig:result_pearson}
\end{figure*}

 \begin{figure*}[t]
    \centering
    \includegraphics[width=16cm]{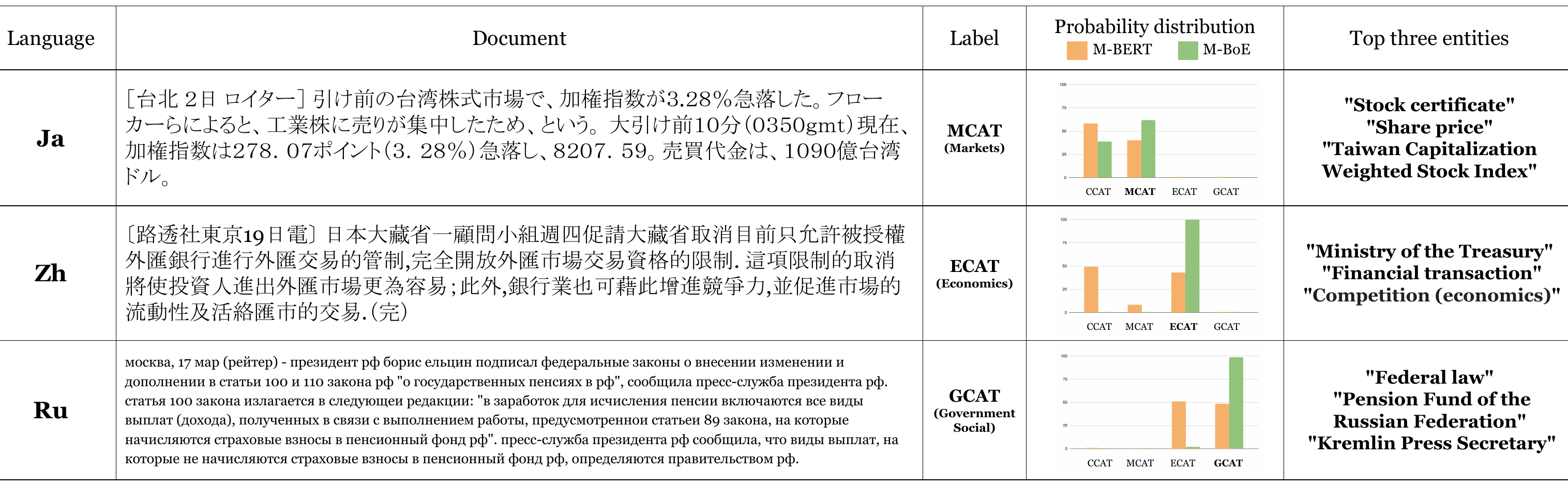}
    \caption{Example results for MLDoc. ``Top three entities'' indicates the three most influential entities selected by attention mechanism.}
    \label{fig:examples}
\end{figure*}

\section{Analysis}

We conducted a series of experiments to analyze the performance of our model on the MLDoc dataset (Table \ref{tab:result_analysis}).
We first analyzed the impact on the performance of each component in the M-BoE model, including the attention mechanism, pre-trained entity embeddings, and entity detection methods.
We then evaluated the sensitivity of the model's performance to differences in the number of detected entities for each language. 
Finally, we conducted qualitative analysis by visualizing important entities.

\subsection{Attention mechanism } 
We examined the effect of the attention mechanism on performance.
When the attention mechanism was removed (Table \ref{tab:result_analysis}: \textbf{Attention mechanism}), the performance was substantially lower than with the proposed model.
This indicates that the attention mechanism selects the entities that are effective in solving the classification task.
Next, we examined the effectiveness of the two features (i.e., cosine and commonness) in the attention mechanism by excluding them one at a time from the M-BoE model. Table \ref{tab:result_analysis} shows that there was a slight drop in performance when either of them was not used, indicating that both features are effective.

\subsection{Entity embeddings } 
To investigate the effect of entity embedding initialization, we replaced Wikipedia2Vec with (1) random vectors and (2) knowledge graph (KG) embeddings (Table \ref{tab:result_analysis}: \textbf{Entity embeddings}).
For KG embedding, we used ComplEx \cite{pmlr-v48-trouillon16}, a state-of-the-art KG embedding method.
We trained the ComplEx embeddings on the wikidata5m dataset \cite{10.1162/tacl_a_00360} using the kge tool.\footnote{\url{https://github.com/uma-pi1/kge}} We set the dimension to 768 and used the default hyper-parameters for everything else in the {\tt wikidata5m-complex configuration} in the tool.
The results show that using Wikipedia2Vec was the most effective although using KG embeddings was better than using random vectors.

\subsection{Entity detection method} 

To verify the effectiveness of our dictionary-based entity detection method, we simply replaced it with a commercial multilingual entity linking system, Google Cloud Natural Language API\footnote{\url{https://cloud.google.com/natural-language}} (Table \ref{tab:result_analysis}: \textbf{Entity detection method}).
All entities were detected with the API and converted into Wikidata entities, as explained in Section \ref{subsec:entity}. Note that unlike our dictionary-based method, the entity linking system detects a single disambiguated entity for each entity name. 

The results show that our entity detection method outperformed the API.
We attribute this to the number of entities detected with our dictionary-based detection method.
As shown in Table \ref{tab:detection_method}, the number of entities detected with the entity linking system was substantially lower than with our entity detection method because, unlike our method, the system detects only disambiguated entities and does not detect non-named entities. 
Therefore, we attribute the better performance of our method compared with that of the API to (1) non-named entities also being important features and (2) the inability to use the correct entity if the disambiguation error is caused by entity linking.

Furthermore, as described in Section \ref{sec: results}, our entity detection method performed competitively with the human-labeled entity annotations on the SHINRA2020-ML dataset.

Next, we examined the performance impact of the number of detected Wikidata entities. 
For the full model and no attention model, we observed a change in performance when some percentage of the entities were randomly removed during training and inference. 
Figure \ref{fig:entity_ratio} shows that, the higher the entity detection rate, the better the performance of the full model.
When the attention mechanism was removed, however, there was no consistent trend. The performance remained the same or even dropped. 
These results suggest that the more entities detected, the better the performance, and that the attention mechanism is important for this consistent improvement.

\subsection{Performance sensitivity to language differences } 

In our method, the number of detected Wikidata entities during inference differs depending on the target languages. We investigated how this affects performance.
For each of the datasets, we computed the Pearson's correlation coefficient between the number of detected entities and the rate of improvement over the baseline performance for each language (Figure \ref{fig:result_pearson}). 
As a result, there was no clear trend in the correlation coefficients, which ranged from -0.3 to 0.2. These results indicate that the performance was consistently improved for languages with a small number of detected entities.
We attribute this to the ability of our method to detect a sufficient number of entities, even for languages with a relatively small number of entity detections.

\subsection{Qualitative analysis}
To further investigate how the M-BoE model improved performance, we took the MLDoc documents that our model classified correctly while M-BERT did not and examined the influential entities that were assigned the largest attention weights by the M-BoE model.
Figure \ref{fig:examples} shows three examples in which the M-BoE model effectively improved performance.
Overall, it identified the entities that were highly relevant to the document.
For example, the first document is a Japanese document about the Taiwanese stock market, and the M-BoE model correctly identified the relevant entities, including \textit{ \textbf{Stock certificate}}, \textit{\textbf{Share price}}, and \textit{\textbf{Taiwan Capitalization Weighted Stock Index}}.


\section{Conclusions}

Our proposed M-BoE model is a simple extension of multilingual PLMs: language-independent Wikidata entities are used as input features for zero-shot cross-lingual text classification.
Since the Wikidata entity embeddings are shared across languages, and the entities associated with a document are further selected by the attention mechanism, a model trained on these features in one language can efficiently be applied to multiple target languages. 
We achieved state-of-the-art results on three cross-lingual text classification tasks, which clearly shows the effectiveness of our method.

As future work, we plan to evaluate our model on low-resource languages and a variety of natural language processing tasks, such as cross-lingual document retrieval. 
We would also like to investigate whether our method can be combined with other methods, such as using additional textual data in the target language. 

\label{sec: conclusion}

\section*{Acknowledgements}

This work was partially supported by JSPS KAKENHI Grants JP16H06302, JP18H04120, JP20K23355, JP21H04907, and JP21K18023, and by JST CREST Grants JPMJCR18A6 and JPMJCR20D3, Japan.


\bibliography{anthology,custom}
\bibliographystyle{acl_natbib}

\newpage
\appendix
\onecolumn

\begin{center}
{\bf \large{
    Appendix for ``A Multilingual Bag-of-Entities Model for  \\ Zero-Shot Cross-Lingual Text Classification''}}
\end{center}

\section{Hyper-parameter Details}
\label{appendix:hyper-parameter_details}

We conduct a grid-search for batch size $\in \{16, 32, 64, 128\}$ and learning
rate $\in \{1e-05, 2e-05, 5e-05\}$. 
The chosen hyperparameters for each model are shown in Table \ref{tab:hyperparameter}.

\begin{table*}[h]
\centering
\scalebox{0.95}{
\begin{tabular}{l|cccc}
\hline
Model & MLDoc & TED-CLDC & SHINRA2020-ML \\
\hline
M-BERT & 32 / 2e-05 & 16 / 2e-05  &  128 / 5e-05 \\

XLM-R & 32 / 2e-05 & 16 / 5e-05  &  64 / 2e-05 \\

M-BoE  (M-BERT) & 32 / 2e-05 & 16 / 2e-05  & 128 /  5e-05\\

M-BoE (XLM-R)& 32 / 2e-05 & 16 / 5e-05  & 64 /  2e-05\\
\hline
\end{tabular}
}
\caption{Hyper-parameters used for experiments. In each cell, the left value indicates batch size, and the right value indicates learning rate.}
\label{tab:hyperparameter}
\end{table*}

\end{document}